\documentclass[review]{elsarticle}
\usepackage{amssymb}
\usepackage{latexsym}
\usepackage{subfigure}
\usepackage{graphicx}
\usepackage{url}
\usepackage{xcolor}
\usepackage{framed,multirow}
\usepackage{bm}
\usepackage{lineno,hyperref}
\usepackage{algorithm}
\usepackage{algorithmic}
\usepackage{setspace}

\journal{Journal of \LaTeX\ Templates}

\begin{document}

\begin{frontmatter}

\title{MM-FSOD: Meta and metric integrated few-shot object detection}


%\author{Yuewen Li\fnref{myfootnote}}
%\address{Beijing, China}
%\fntext[myfootnote]{Since 2019.}

\author[mymainaddress]{Yuewen Li}
\ead[url]{syliyuewen@buaa.edu.cn}
\author[mymainaddress]{Wenquan Feng}
\ead[url]{buaafwq@buaa.edu.cn}
\author[mymainaddress]{Shuchang Lyu}
\ead[url]{lyushuchang@buaa.edu.cn}
\author[mymainaddress]{Qi Zhao}
\ead[url]{zhaoqi@buaa.edu.cn}

\author[mymainaddress]{Xuliang Li\corref{mycorrespondingauthor}}
\cortext[mycorrespondingauthor]{Corresponding author}
\ead{xulli8997@buaa.edu.cn}

\address[mymainaddress]{Beihang University, Beijing, China}

\begin{abstract}
In the object detection task, CNN (Convolutional neural networks) models always need a large amount of annotated examples in the training process. To reduce the dependency of expensive annotations, few-shot object detection has become an increasing research focus. In this paper, we present an effective object detection framework (MM-FSOD) that integrates metric learning and meta-learning to tackle the few-shot object detection task. Our model is a class-agnostic detection model that can accurately recognize new categories, which are not appearing in training samples. Specifically, to fast learn the features of new categories without a fine-tuning process, we propose a meta-representation module (MR module) to learn intra-class mean prototypes. MR module is trained with a meta-learning method to obtain the ability to reconstruct high-level features. To further conduct similarity of features between support prototype with query RoIs features, we propose a Pearson metric module (PR module) which serves as a classifier. Compared to the previous commonly used metric method, cosine distance metric. PR module enables the model to align features into discriminative embedding space. We conduct extensive experiments on benchmark datasets FSOD, MS COCO, and PASCAL VOC to demonstrate the feasibility and efficiency of our model. Comparing with the previous method, MM-FSOD achieves state-of-the-art (SOTA) results.
\end{abstract}

\begin{keyword}
Few-shot object detection\sep Meta Learning\sep Metric learning\sep Feature representation\sep Pearson distance
\MSC[2010] 00-01\sep  99-00
\end{keyword}

\end{frontmatter}

\section{Introduction}
\label{sec1}
Existing computer vision tasks always rely on a large amount of annotated data to train models. However, collecting and annotating large amounts of training data costs a lot. To reduce this cost, few-shot learning is introduced. Few-shot learning aims to enable CNN networks to effectively recognize new targets with only a few samples. In recent years, few-shot learning has achieved significant progress in image classification, segmentation and objection detection fields~\citep{LeeKetal2019,RenMetal2019, Lyuetal2019, Perez-Ruaetal2020, Fanetal2020}. Few-shot learning can roughly be categorized into two categories: metric learning~\citep{ZhangCetal2020,BrattoliBetal2020, BateniPetal2020} and meta learning~\citep{WangYetal2020,SachinRavietal2017}. Metric learning focuses on creating an embedding space that can significantly distinguish different categories by designing metric distance and loss functions. As for meta-learning, there are several methods, such as training modules with model-agnostic method~\citep{Finnetal2017}, and these modules are called meta-learners. Meta-learner recognizes accurately new categories in the test dataset by learning from a small amount of annotated data and obtaining a set of parameters that are sensitive to those categories.
Few-shot classification task aims to improve the accuracy of classification by using meta-learning and metric learning~\citep{SunQetal2019, LeeKetal2019}(add more). Most few-shot works focus on classification rarely going into the task of few-shot object detection, most probably because transferring from few-shot classification to few-shot objection is a non-trivial task. The focus of few-shot object detection is how to localize new classes of objects in a complex background. Some few-shot object detection methods~\citep{Kangetal2019, Yanetal2019, Fanetal2020} apply meta-learning method to recognize new class objects with few training examples. Methods such as ~\citep{Yanetal2019, Kangetal2019} train a meta-feature extractor to encode the features of new categories and integrate them into query images that need to be recognized. These methods enhance the expression of the new class, but cannot be directly applied to novel categories. Furthermore, most few-shot object detection works focus on how to enhance the expression in deep space. They lose sight of considering the distance of inter-class and intra-class in deep space. When it comes to multiple class detection, it is important to rethink how to express different class features with a maximal distance from each other. As far as we know, metric learning is rarely applied in few-shot object detection. Because the performance of the model is limited by the properties of the distance function. Cosine distance is the most popular method in metric learning, but there are still flaws. For cosine distance, it requires that features in predicted vectors are independent of each other, which will make desired results. However, in most cases, these features are coupled, which is fatal for cosine distance.

 \par Motivated by the above problems, we propose an end-to-end trainable Few-shot object detection model named MM-FSOD, which incorporates meta-learning and metric learning. Inspired by ~\citep{Karlinskyetal2019, Finnetal2017}, we adopt an episode-based approach to train the meta-learning module, which enables this module to fast converge in a class-agnostic paradigm. Also, due to the standard criteria of metric learning ~\citep{Snelletal2017}, we use the distance formula to replace the traditional classification layer as the final classification criteria. By combining the two approaches, we explore a method for minimizing over-fitting effects and enhancing the generalization of CNN models, as well as achieving fast convergence to new classes.

 \par We all know that a few annotated examples of the new class will make it difficult for CNN models to learn useful knowledge. In other words, the model is more sensitive to fully trained categories~\citep{Zhaoetal2019}. The insensitivity to the new category targets results in poor representation in the deep feature space. To address this issue, some class-specific methods like~\citep{Kangetal2019, Yanetal2019} fuse the features of support examples into the query examples by channel-wise multiplication. Different from class-specific methods, our method aims to design a class-agnostic object detection model that can detect targets of new classes without additional offline learning. To achieve this, we design a meta representation(MR) module to reconstruct the low-level features in the deep network with the meta-learning method. Different from the meta-learner in previous methods, which are usually used as a classifier, the MR module represents low-level features in deep space and expands the inter-class distance.

 \par To classify targets with metric learning, many methods~\citep{LiuJetal2019,Mai2019} use common cosine distance to compute the similarity between support objects and query objects. However, using cosine distance in object detection task has some problems. The complex background information in query images will cause a large intra-class variance. And the features in the predicted vector after CNN network is Mutually coupled, which will make cosine distance give a wrong result. To solve this problem, we use the Pearson distance to alleviate the influence of intra-class variance and uncouple features in predicted vectors. Pearson distance normalizes feature vectors before calculating similarity, which reduces the effect of intra-class variance and takes into account the coupling between features of predicted vectors. Additionally, the MR module learns from the intra-class mean prototypes of support examples of different categories. And using Pearson distance can force the features to cluster around the mean prototypes. Therefore, We use Pearson distance, which is more suitable than cosine distance for the MR module.

\par We conduct extensive experiments to demonstrate that our method performs better than previous methods. On few-shot detection benchmark datasets FSOD~\citep{Fanetal2020}, PASCAL VOC~\citep{Everinghametal2010} and MS COCO ~\citep{Linetal2014}, our method achieve new state-of-the-art (SOTA) results on most datasets. Furthermore, we use ablation studies to prove that combining the Pearson distance and MR module performs better.

\par The contribution of our model can be summarized as follows:
\begin{itemize}
\item We propose an end-to end few-shot object detection model that effectively integrate meta learning and metric learning methods.
\item We design a feature reconstruction module named MR module that reconstructs low-level features to clearly distinguish the clustering centers of different classes.
\item We introduce Pearson distance in metric learning and illustrate its contribution for tackling coupling features and intra-class variance.
\item We conduct experiments on benchmark datasets to show that MM-FSOD achieves SOTA results compared to previous methods.
\end{itemize}

\begin{figure*}[htbp]
  \centering
  \includegraphics[width=12cm, height=4cm]{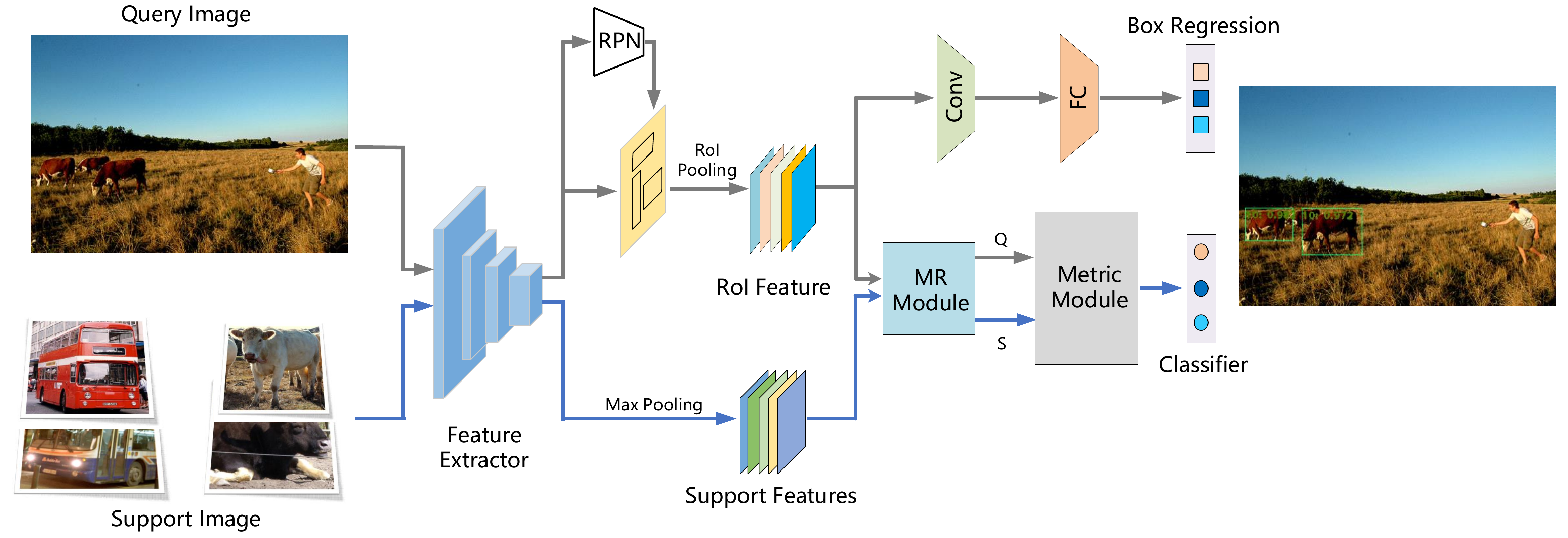}
  \caption{The workflow of MM-FSOD: Our MM-FSOD is based on Faster R-CNN. The Feature Extractor receives an query image to generate RoI features by taking  RoI-Pooling on image region proposals extracted by RPN. For support set, we just take the extracted features after feature extractor and do max-pooling on them. Then we send query features and support features into MR module and Metric Module for classification process. In parallel, we send query RoI features into bounding box prediction branch to obtain final object location.}
  \label{Fig.1}
\end{figure*}

\section{Related work}
{\bf Object Detection.} Object detection~\citep{Caietal2018, Kimetal2018} is a classical computer vision task. Recent object detection architectures are divided into two main types: one-stage detectors and two-stage detectors. From another perspective, object detectors can be divided into anchor based and anchor free detector. One-stage architectures like YOLO~\citep{Redmonetal2016, Redmonetal2018, Zhouetal2019} series use CNN network to directly predict bounding boxes and classification. Two-stage architectures, for example, Faster R-CNN~\citep{Renetal2015,Heetal2017}, first generates class-agnostic region proposals of potential objects from a given image. And then these proposals are sent into the second stage CNN networks to refine the bounding box and classified. This method uses the RPN module to overcome the influence from many negative locations which can help the next classifier achieving accurate classification. Benefiting from RPN, two-stage methods usually perform better than one-stage methods in object detection task. Our model is based on Faster R-CNN, because its second stage can be easily designed for few-shot object detection.

\par {\bf Few-Shot Learning.} In terms of methodology, there are currently two main effective approaches: meta-learning and metric learning. Metric learning~\citep{BateniPetal2020, LiWetal2019, LiuJetal2019} focuses on the design of distance formulation to distinguish different categories. These methods train models to map images into an embedding space and then measure the distance of support and query features by using cosine distance. TADAM~\citep{Oreshkin2018} finds that using metric scaling which multiplies a learnable factor for each scale distance can obtain promising results. Meta learning~\citep{Munkhdalaietal2018, MishraNikhiletal2017, WangYetal2020, SunQetal2019, LeeKetal2019, Sungetal2018} aims to train network parameters with only a few examples of new categories. These methods train a meta learner to learn from the distribution of learning tasks and then acquire generalization capability. MAML~\citep{Finnetal2017} is a gradient-based method that uses gradient descent to adapt the embedding model parameters with few training examples. Similar to MAML, other methods construct a meta learner, which consists of fully connected networks~\citep{SunQetal2019} or linear classifiers~\citep{LeeKetal2019}. Optimized through updating parameters in several steps, the meta learner can adapt to new tasks quickly. Some approaches~\citep{RenMetal2019, MishraNikhiletal2017} introduce attention mechanisms into few-shot learning to enhance the deep network's feature expression.

\par {\bf Few-Shot Object Detection} Few-shot learning task is different from traditional computer vision tasks because only a few examples are available for training, which is difficult for most machine learning algorithms. Recent works~\citep{Dongetal2018, Wang2020} have attempted to apply few-shot learning to objection detection. The feature reweighting method~\citep{Kangetal2019} uses YOLO~\citep{Redmonetal2018} and a meta-learner module to extract representative features of support examples then apply them to the query features in a channel-wise manner. Similarly, Meta R-CNN~\citep{Yanetal2019} designs a PRN network to extract attention vectors of support examples and fuse them with query RoI features in a depth-wise manner. Different from the above methods by combining the support feature vectors to the query features, ONCE~\citep{Perez-Ruaetal2020} first proposes an incremental-learning based few-shot object detection method, which directly learns support examples to predict a set of detector parameters. Unlike class-specific architectures that need to fine-tune on new classes, FSOD~\citep{Fanetal2020} proposes a new class-agnostic few-shot detection network. The core of the approach is attentive RPN, multi-relational detector, and comparative training strategy. It leverages the similarity between support and query examples to detect new objects. Before the multi-relational detection module, the feature vectors of support samples are multiplied with features of query examples in a depth-wise manner before feeding into prediction layers. Chen et al. ~\citep{ChenHetal2018} proposes a novel few-shot method that employs a new regularization method to enhance fine-tuning. The network consists of transfer knowledge (TK) and background suppression (BD). TK transfers source domain knowledge of each target to the target domain to generalize few-shot learning. BD uses the bounding box knowledge of the target image as additional supervision of the feature map. The background interference is reduced when focusing on the targets.

\section{Proposed method}
In this section, we first show our few-shot object detection model and then specifically describe the novelty of our proposed model.
\subsection{Problem definition}
In few-shot learning field, we respectively denote support images and query images as $S_{c}$ and $Q_{c}$. The goal of few-shot object detection is to detect new class targets in query image with the help of support images and its annotations. The support set contains \emph{N} classes and \emph{K} examples for each class, which is called N-way K-shot task. We commonly divide the whole categories of datasets into two parts ($C_{base}$ and $C_{novel}$). $C_{base}$ is used for training and $C_{novel}$ is used for testing. The intersection of these two parts is empty $C_{base} \cap C_{novel} = \varnothing $.
\subsection{The architecture of MM-FSOD}
The overview of our model is shown in Fig.~\ref{Fig.1}. Our few-shot object detection model is inspired by Faster R-CNN framework. We design several novel modules to better tackle few-shot object detection task. As shown in Fig.~\ref{Fig.1}, the support and query image share the weights of the feature extractor. The same backbone ensures that the low-level features of support and query images are lying in the same embedding space. After feature extraction, features of support and query images respectively pass through max-pooling layer and RPN module to obtain support features and RoI features. Using max-pooling on support image features is to highlight obvious semantic information. Features of query image are fed into RPN module to obtain the RoI features with foreground and background information. As for query features, we conduct RoI-Pooling. Then we send the support features and query RoIs features into the MR module for feature reconstruction.
\par In MR module, we use support features for iterative learning. We extract the intra-class mean prototypes of each category and feed them into the fully connected layer for classification. After a certain step of learning and continuous optimization of MR module, the query RoIs features are sent to MR module to reconstruct features. Then, we replace the FC layer with support intra-class prototypes.
\par  In metric module, we use Pearson distance to obtain similarity between each RoI and prototype. Compared to popular cosine distance, Pearson distance is more suitable for features represented by MR module. Because, Pearson distance reduces the negative effect from large variance and coupling relationship of features.
\par For the location prediction, we keep the basic/bottleneck blocks (layer4) and FC layer of ResNet to predict localization offsets in a class-agnostic manner. We keep the box branch unchanged, because the classification of box is unnecessary. We only need to know that it is a foreground region. In our experiments, we found that while the labeling information for each category is sparse for few-shot object detection, the number of labeled borders for foregrounds are relatively high.

\begin{figure}[htp]
  \centering
  \includegraphics[scale=0.73]{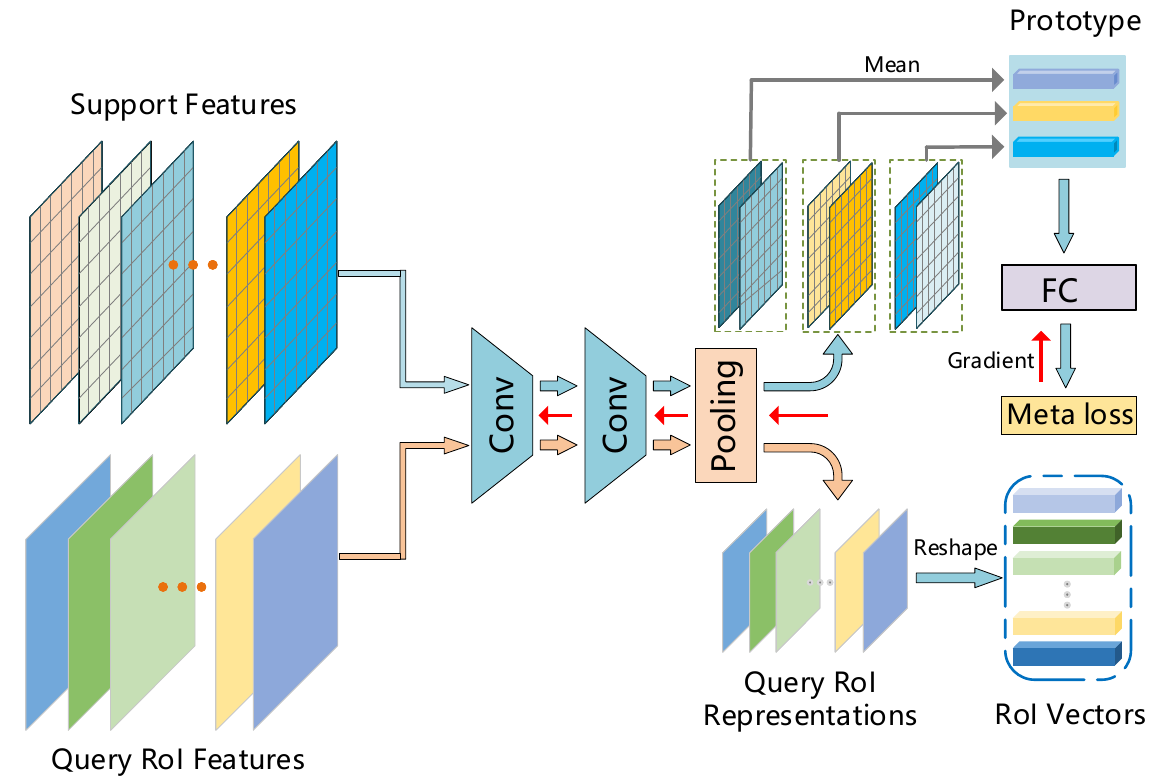}
  \caption{This diagram shows how the module works and its' elements. The support features is refined by a lightweight convolution network and compute the intra-class means as class prototypes for meta learning. The red arrows representing the backward path of the gradient of support prototypes. After some steps learning from support prototypes, we use MR module to reconstruct the query RoI features.}
  \label{Fig.2}
\end{figure}

\subsection{Meta-representation module}
 As shown in  Fig.~\ref{Fig.2}, the image is down-sampled 16 times by backbone and the output features ($\bm{F}^{\frac{w}{16} \times \frac{h}{16} \times d}$) are fed into MR module. We formulate the feature extraction process of backbone in Eq.~\ref{eq1}

\begin{equation}
\bm{F} = g(\bm{X} ; \bm{W}),\quad \bm{F} \in R^{\frac{w}{16} \times \frac{h}{16} \times d},\quad \bm{X} \in S_{c} \cup Q_{c}
\label{eq1}
\end{equation}
where \emph{g($\cdot$)} is the backbone of the model, and \emph{d} is the number of channels.

\par In order to enhance the expression of low-level features, MR module is designed to perform feature reconstruction, which consists of a lightweight CNN module that contains two ${3 \times 3}$
convolution layers, a pooling layer and a fully connected layer. The first convolution kernel of MR module performs on $\bm{F}$ with a stride of 2 to obtain $\bm{F}^{\frac{w}{32} \times \frac{h}{32}\times2d}$ features. The second convolution kernel has a stride of 1 to refine the features. After the features are reconstructed by two convolution kernels, we adopt average pooling and reshape operations to obtain the overall semantic information of input features, which can combine the overall feature information and conducts more representative features. In particular, suppose that MR module denotes $h(\cdot; \bm{\theta})$, given each support featuer and query RoI feature($\mathbf{F}$), it holds:
\begin{equation}
\bm{v_{meta}} = h(g(\bm{X} ;~\bm{W}); \mathbf{\theta}), \quad \bm{X} \in S_{c} \cup Q_{c}
\label{eq2}
\end{equation}

where $\bm{\theta}$ denotes the parameters of MR module.

\begin{figure*}[t]
  \centering
  \includegraphics[scale=0.8]{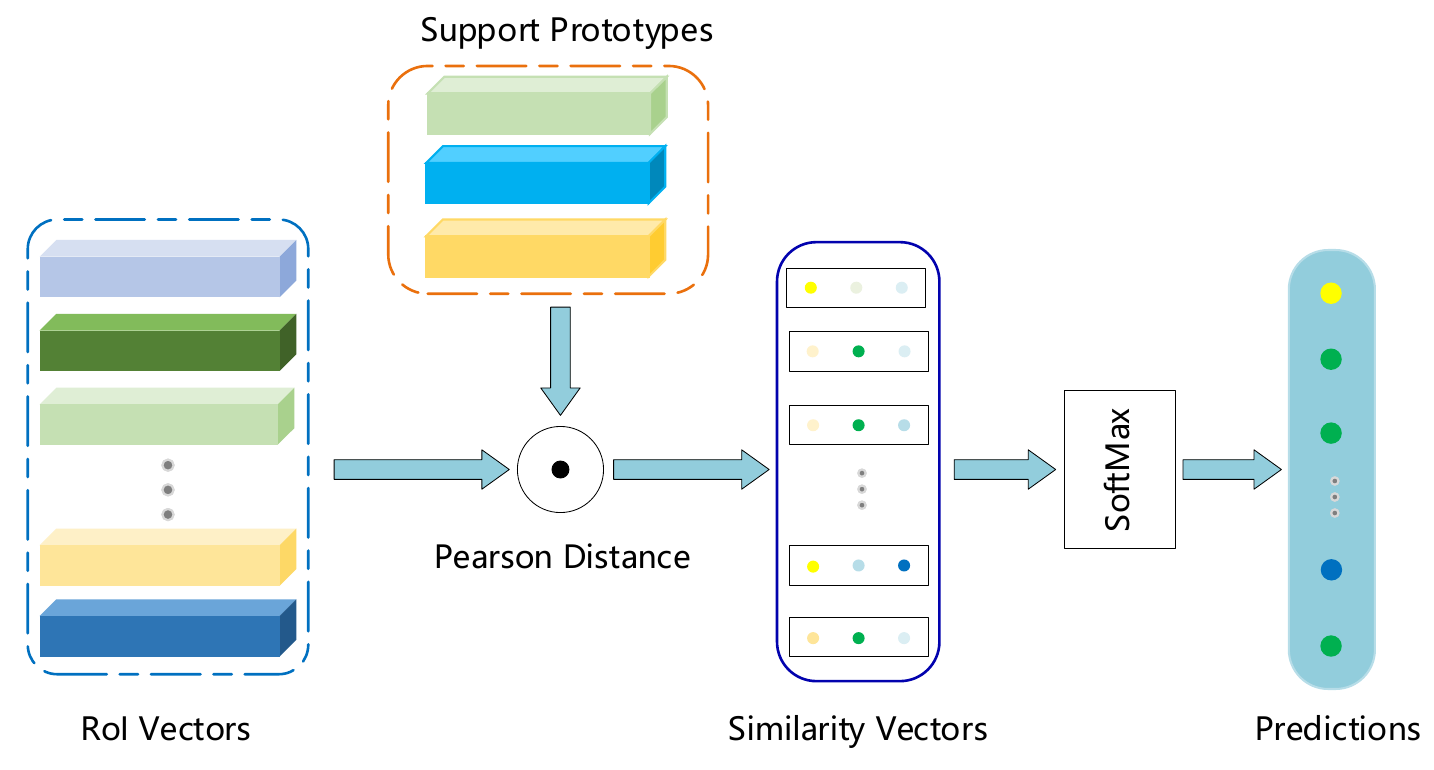}
  \caption{Metric module. The figure shows a 3-way 5-shot process. There are three prototypes in the diagram, one of which is the background class prototype.}
  \label{Fig3}
\end{figure*}

\par The above is a general method to process query and support features in convolutional neural network. Usually, training a CNN network needs specific label for each class. However, instead of assigning a fixed label to each support images during training, we use a method similar to prototype networks~\citep{LiuJetal2019} to compute the intra-class mean of support features and obtain a 1-dimension representative prototype vector for each class:
\begin{equation}
\bm{s_{c_n}} = \frac{1}{k}\sum_{j=1}^k \bm{v_{meta}}(c_n, j), \quad c_n \in C,\quad k=1,2,3\cdots
\label{eq3}
\end{equation}

where \emph{C} is the training class set and $s_{c_n}$ is the prototype of class $c_n$. Then each prototype vector is assigned random label before feeding into the fully connected layer for classification. The reason for assigning random label is that we want MR module to learn how to distinguish representative features of different classes and how to expand the distance of different classes' feature in feature space. By doing this, the module actually learns how to distinguish the currently learned classes in a class-agnostic manner.

\begin{algorithm}[t]
\setstretch{1.35} %设置具有指定弹力的橡皮长度（原行宽的1.35倍）
\caption{}
\label{alg1}
\begin{algorithmic}
\REQUIRE ~~\\
${\emph{p(T)}}$: distribution of all tasks;\\
${\alpha}$, ${\beta}$: meta learning rate and step size;\\
randomly initialize $\theta$ \\
Sample episode of tasks $T_i$ ${\sim}$ \emph{p(T)} \\
\FOR{all $T_i$}
\STATE Compute the intra-class mean prototype $S_{c_{i}}$ of support set for each class
\STATE Evaluate loss with respect to optimise parameters with gradient decent:\\
$\theta_{i}^{'}$ = $\theta$ - lr $\bigtriangledown_{\theta}L_{T_{i}}(f_{\theta})$
\ENDFOR
\STATE reconstruct the query RoIs features
\end{algorithmic}
\end{algorithm}

\par For optimization of this module, we adopt a model-agnostic~\citep{Finnetal2017} method to learn how to classify support prototypes accurately. We use cross-entropy loss and conduct gradient descent to optimize MR module:
\begin{equation}
\theta_{i}^{'} = \theta - lr \bigtriangledown_{\theta}L_{T_{i}}(f_{\theta})
\label{eq4}
\end{equation}
where \emph{lr} is the learning rate in the meta-learning phase. \emph{T$_{i}$} is the learning task sampled from total tasks. We set a certain number of optimization step, so that the module can fully learn how to reconstruct the support features and expand the distribution distance of different category prototypes. After learning experience knowledge from the support images, MR module obtains the ability to conduct features of the same category into close distribution regions. Then we use the optimized MR module to reconstruct the features of target RoIs in query images. Instead of averaging the reconstructed features and feeding them into the FC layer like the process of support images. We stretch the average pooled query RoIs features into 1-dimension vectors same as the support prototypes. Finally, we feed reconstructed features into metric module and implement the specific classification criteria. The process of how to train the MR module can be seen in Algorithm.\ref{alg1}.

\subsection{Pearson distance}
 \par Fig.~\ref{Fig3} shows the workflow of metric module. The query RoIs features come from MR module and are reshaped into 1-dimension vectors. Then the similarity vectors are calculating by Pearson distance operation between the support prototypes and RoI vectors. We classify query RoIs according to the principle of closest distance.
 \par In metric module, we mainly introduce two types of distance, cosine distance and Pearson distance. For the distance function, we learn from past work~\citep{KochGetal2015} that Euclidean distance and KL distance are not performed well on few-shot learning. Previous works prove that cosine distance has good performance in metric learning~\citep{LiWetal2019,Mai2019}.

\begin{equation}
D(\bm{v^{roi}_{meta}}, \bm{s_{c_n}}) = \frac{\bm{v^{roi}_{meta}} \cdot \bm{s_{c_n}}}{\parallel{\bm{v^{roi}_{meta}}} \parallel{} \cdot  \parallel{\bm{s_{c_n}}} \parallel{}} , \quad 1 < n < N
\label{eq5}
\end{equation}

where $\bm{s_{c_n}}$ denotes the prototype vector of class \emph{n} and $\bm{v^{roi}_{meta}}$ is the reconstructed feature vector of RoI. $D(\bm{v^{roi}_{meta}},~\bm{s_{c_n}})$ is the distance between support prototype and RoI vector with a distribution ranging from 0 to 1. 0 represents the maximum distance and 1 represents the closest distance.

\begin{figure*}[t]
  \centering
\subfigure[]{
\includegraphics[width=5.5cm]{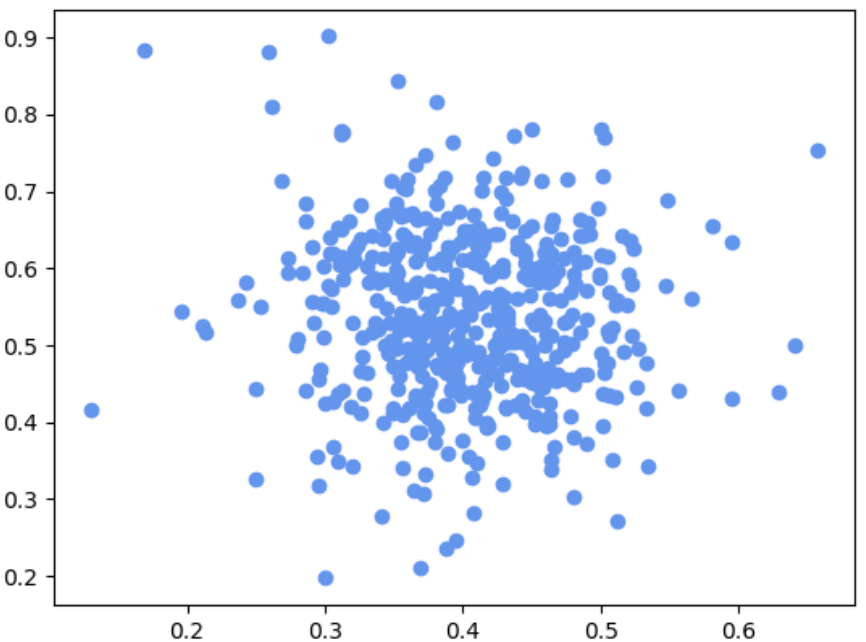}
\label{Fig5.a}
%\caption{fig1}
}
\quad
\subfigure[]{
\includegraphics[width=5.5cm]{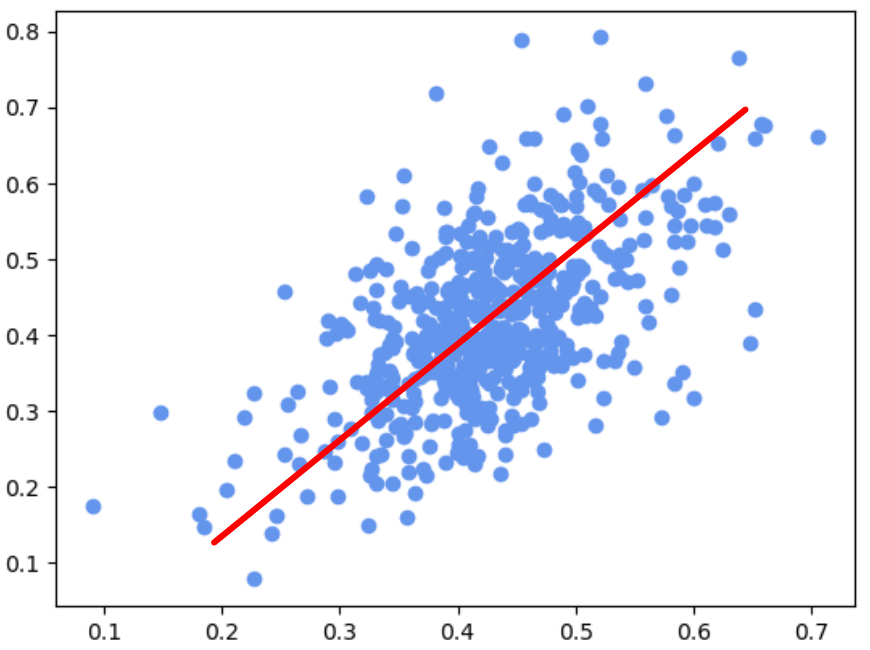}
\label{Fig5.b}
}
  \caption{The relationship between query RoIs and support prototype under Pearson distance. If query RoIs and support prototypes do not have a linear relationship}
  \label{Fig5}
\end{figure*}

\par Different from cosine distance, Pearson distance calculates similarity between support prototypes and RoI vectors by normalizing each dimension with the mean of all dimensions:
\begin{small}
\begin{equation}
PR{(\bm{v^{roi}_{meta}}, \bm{s_{c_n}})} = \frac{\sum_{i=1}^d(\bm{v^{roi}_{meta}}(i) - \overline{\bm{v^{roi}_{meta}})}(\bm{s_{c_n}}(i) - \overline{\bm{s_{c_n}}})}{\parallel{\bm{v^{roi}_{meta}}(i) - \overline{\bm{v^{roi}_{meta}}}} \parallel{} \cdot  \parallel{\bm{s_{c_n}}(i) - \overline{\bm{s_{c_n}}}} \parallel{}}
\label{eq6}
\end{equation}
\end{small}
where $PR{(\bm{v}^{roi}_{meta},~\bm{s_{c_n}})}$, $\bm{s_{c_{n}}}$ represents the similarity between each RoI vector and each support prototype. \emph{d} means the number of dimensions of query RoI vector and support prototype vector.  $\overline{\bm{v}^{roi}_{meta}}$ and $\overline{\bm{s}_{c_n}}$ are the mean of their vectors overall dimensions. From Fig.\ref{Fig5} we can see that when targets in query images are not correlated with support prototypes, their relationship is not linear(Fig.~\ref{Fig5.a}). When they are strongly correlated, it will show a linear relationship as Fig.~\ref{Fig5.b} shows.
\par Although cosine distance is good enough to evaluate the distance between two different vectors, it is not the best choice for our model. The reason is that there is usually a lot of background regions in RoIs features extracted by RPN module. Some of these background regions are intermediate targets that just are not current learning categories. This part of the intermediate targets bothers cosine distance a lot. Because the background class in support set is randomly cropped from annotated images, and intermediate targets may have some similarity to foreground classes. It will often cause the cosine distance to give wrong similarity scores that greater than 0.5 for intermediate targets. As each learning task is randomly sampled from datasets, there will be inescapable to sample a part of learning tasks that contains similar categories in query images. When this always happens, cosine distance will lose the ability to help the model to learn an embedding space.

\begin{figure}[t]
  \centering
  \includegraphics[scale=0.65]{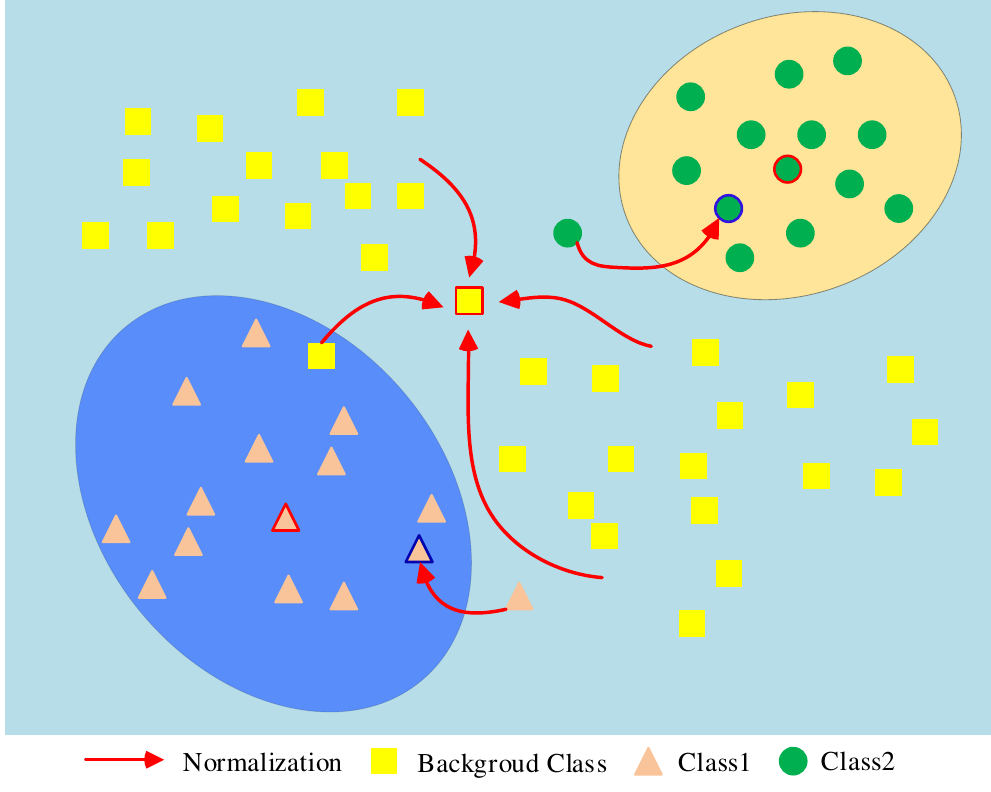}
  \caption{The working mechanism of Pearson distance in metric module. Before calculating the distance between each RoI and support prototype, the reconstructed features of RoI will be normalized towards their prototypes.}
  \label{Fig4}
\end{figure}

\par Complex background information not only affects cosine distance but also causes the MR module to deviate reconstructed RoIs features from the optimal cluster region of their real class prototype. This will make some RoIs features located on edge of different classes cluster regions. It is hard for cosine distance to classify these targets on the boundary region. To solve these problems, Pearson distance Eq.~(\ref{eq6}) normalizes each dimension of input features before calculating similarity. This leads the features of each dimension to move closer to the clustering center, then the distance from the other categories is also increased as showing in Fig~\ref{Fig4}. This method reduces the misjudgment of the intermediate targets on the boundary region.
\par After calculating the similarity of RoI vectors and support prototypes as showing in Fig.~\ref{Fig3}, we use the softmax function to normalize the similarity vector of each class. In the training process, we find that if the similarity vectors are fed directly into softmax will cause low confidence in output. By the way, if the score threshold is set too high, the positive RoIs will be neglected. To solve this problem, we follow{~\citep{LiuJetal2019}}  and set a hyperparameter ${\alpha}$ = 10 to expand the output value of the similarity:
\begin{small}
\begin{equation}
P(\bm{v^{roi}_{meta}}, c_{n}) = \frac{e^{PR{(\bm{v^{roi}_{meta}},~\bm{s_{c_{n}}}) \cdot \alpha}}}{\sum_{n=1}^N e^{PR{(\bm{v^{roi}_{meta}},~ \bm{s_{c_{n}}})} \cdot \alpha}}, \quad c_n \in C,\quad n=1,2,3\cdots
\label{eq7}
\end{equation}
\end{small}
where $P(\bm{v^{roi}_{meta}}, c_{n})$ represents the confidence score of RoI to each class. Finally, we obtain a classification result according to the subscript where the maximum confidence is located.

\subsection{Loss function}
MR module has the ability to reconstruct features and expand the distances between different categories. Mean while, Pearson distance plays a similar role to FC layer. So, there is no need to design a special distance loss function that helps the model learn an embedding space. We use cross-entropy loss to calculate the classification loss:
\begin{equation}
    L_{cls} = - \sum_{n=1}^N{\hat{y_{n}}ln P(\bm{v^{roi}_{meta}}, c_{n})}
    \label{eq8}
\end{equation}
where \emph{N} is the number of training classes in support set, $\hat{y_{j}}$ denotes the real label of target in query images.
\par Unlike MAML~\citep{Garciaetal2017}, the loss and gradient of support set are not only used for the MR module in our method. When classifying the query RoIs, we still need to perform a single forward computation on support set in addition to obtain the support prototypes as a classification criterion. Therefore, the classification gradient is actually computed with both the gradient of the query RoIs and support prototypes. We compute the gradient backward to the MR module separately to obtain:
\begin{small}
\begin{equation}
    \frac{\partial L_{cls}}{\partial \bm{v^{roi}_{meta}}(i)} = \frac{\partial L_{cls}}{\partial PR{(\bm{v^{roi}_{meta}}, \bm{s_{c_n}})}} \cdot \frac{\partial PR{(\bm{v^{roi}_{meta}}, \bm{s_{c_n}})}}{\partial \bm{v^{roi}_{meta}}(i))}
    \label{eq10}
\end{equation}
\end{small}
\begin{equation}
    \frac{\partial L_{cls}}{\partial PR{(\bm{v^{roi}_{meta}}(i), \bm{s_{c_n)}}}} = \frac{e^{PR{(\bm{v^{roi}_{meta}},~\bm{s_{c_n})}}}}{ e^{\sum_{k}^N{PR{(\bm{v^{roi}_{meta}},~\bm{s_{c_n})}}}}} \sum_{n}^N \bm{y_n} - \hat{\bm{y_{n}}}
\label{eq11}
\end{equation}
\begin{small}
\begin{equation}
    \frac{\partial PR{(\bm{v^{roi}_{meta}}, \bm{s_{c_n})}}}{\partial \bm{v^{roi}_{meta}}(i)} = \frac{(\frac{1}{d} - 1 )\sum_{i=1}^d(\bm{v^{roi}_{meta}}(i) - \overline{\bm{v^{roi}_{meta}}})(\bm{s_{c_n}}(i) - \overline{\bm{s_{c_n}}})}{(\sum_{i}^d(\bm{v^{roi}_{meta}}(i) - \overline{\bm{v^{roi}_{meta}}})^2)^{\frac{3}{2}} (\sum_{i}^d (\bm{s_{c_n}}(i) - \overline{\bm{s_{c_n}}})^2)^{\frac{1}{2}}}
\label{eq12}
\end{equation}
\end{small}
\begin{small}
\begin{equation}
    \frac{\partial PR{(\bm{v}^{roi}_{meta}, \bm{s_{c_n}})}}{\partial \bm{s_{c_n}}(i)} = \frac{(\frac{1}{d} - 1 )\sum_{i=1}^d(\bm{v^{roi}_{meta}}(i) - \overline{\bm{v^{roi}_{meta}}})(\bm{s_{c_n}}(i) - \overline{\bm{s_{c_n}}})}{(\sum_{i}^d (\bm{v^{roi}_{meta}}(i) - \overline{\bm{v^{roi}_{meta}}})^2)^{\frac{1}{2}} (\sum_{i}^d (\bm{s_{c_n}}(i) - \overline{\bm{s_{c_n}}})^2)^{\frac{3}{2}}}
\label{eq13}
\end{equation}
\end{small}

where \emph{d} is the number of channels of the feature vector and \emph{i} represents the \emph{i-th} dimension. $\frac{\partial PR{(\bm{v^{roi}_{meta}}, \bm{s_{c_n})}}}{\partial \bm{v^{roi}_{meta}}(i)}$ represents the gradient of the query set for the MR module and $\frac{\partial PR{(\bm{v^{roi}_{meta}}, \bm{s_{c_n})}}}{\partial \bm{s_{c_n}}(i)}$ represents the gradient of the support prototypes for the MR module.

\par Eq.(\ref{eq12}) and Eq.(\ref{eq13}) show that our method is significantly different from the model-agnostic training methods, which are based on the MAML's idea when it comes to gradient back-propagation. The MAML based training methods do not back-propagate the gradient of support set, which only uses the gradient of support set to optimize the meta-leaner. In our method, the loss of support set not only optimizes the MR module but also optimizes the parameters of whole model. It seems to us that the loss of experience is also needed for the overall parameters of whole model, which plays a role in the fast convergence. According to the gradient back-propagation procession that contains two kinds of gradients, our approach is an end-to-end model that incorporates both meta-learning and metric learning. Not only a discriminative embedding space is trained by metric learning method but also trains an MR module that learns to reconstruct low-level features by meta-learning.
\par In bounding boxes prediction, we take a class-agnostic prediction approach, and only foreground regions with a confidence score exceeding 0.7 need to be marginalized for regression. Therefore, we use the smooth L1 loss to measure the accuracy of the location prediction:

\begin{equation}
    L_{reg} = \sum_{i \in (x,y,w,h)} smooth_{L1}(t_i - t_{i}^*)
    \label{eq14}
\end{equation}

where $t_{i}^{*}$ means the true value and $t_i$ is the prediction.
So the overall loss of the model is as follows:
\begin{equation}
    L_{det} = L_{cls} + \lambda L_{reg}
\label{eq16}
\end{equation}

\section{Experiments}
In this section, we evaluate our model and compare it with various baseline to show that our model is more powerful on different object detection datasets. We mainly evaluate our method on three benchmark datasets:
FSOD~\citep{Fanetal2020}, PASCAL VOC~\citep{Everinghametal2010} and MS COCO~\citep{Linetal2014} .

\subsection{Training strategy}
We adopt an episode based training method similar to ~\citep{SunQetal2019}. Each episode consists of N-way ${\times}$ K-shot examples, where N is the number of training classes and K is the training examples per class. The data of each episode is divided into support set and query set. Support set is used to train the MR module and query set is used to optimize the overall model guiding the model to learn more generalized knowledge. We random sample every episode from the training dataset. As for background class, we crop none bounding box annotated area as background class.
\subsection{Implementation details}
To compare with FSOD~\citep{Fanetal2020}, we let N=3 (2 foreground class and 1 background class) and K is set to 5 for 5-shot training. The number of query images for each class is 6 in training process. The initial learning rate for whole model is set to 0.001 with the learning rate decay strategy. Meta learning rate is 0.01 and the inner training step is 30 for MR module in meta learning-manner. We train MM-FSOD with ResNet34 for 300 epochs and each epoch contains 200 iterations. The step of learning rate decay is set to 48000. Our method is implemented on Pytorch with NVIDIA GTX 1080ti. Our code will be soon available online.

\subsection{Datasets}
\subsubsection{PASCAL VOC}
We follow the settings of~\citep{Kangetal2019}. As VOC has 20 categories, we construct three splits similar to~\citep{Wuetal2020, Kangetal2019, Yanetal2019} and each split has 15 base classes and 5 novel classes. During trianing, we set K$_{sppport}$  = 5 and the N = 2. Only base classes are trained and we do not fine-tune on novel categories like class-specific method~\citep{Kangetal2019, Yanetal2019}. For more details of these three splits, Novel-category split 1(bird, bus, cow, mbike, sofa / rest); Novel-category split 2(aero, bottle, cow, horse, sofa / rest); Novel-category split 3(boat, cat, mbike, sheep, sofa / rest).
\subsubsection{FSOD}
FSOD is a new few-shot object detection dataset created by~\citep{Fanetal2020}. This dataset contains 1000 categories with more than 60k images and 182k bounding boxes in total. Each category contains around 100 samples and about 500 categories have less than 100 samples.

\subsubsection{MS COCO datasets}
MS COCO dataset~\citep{Linetal2014} contains 80 classes, which is collected from various scenes on Flickr. The datasets have three splits for instance detection, Image Caption and object key-point tasks. We follow the settings of ~\citep{Wuetal2020, Kangetal2019} which selects 20 classes of images (same categories as VOC dataset) to form test dataset and the remaining 60 classes of images to form train dataset. When training, we use train-val set to train model and test on test set for 20 novel categories. Besides, We adopt the same image augmentation as training on FSOD.
\subsection{Results}
For each dataset, the best performance is highlighted in tables. Since our method is class-agnostic, we sample 1000 episodes for five times and calculate the average of five results as final result. We adopt standard metrics, such as AP, AP$_{50}$ and AP$_{75}$ to evaluate our model.

\begin{table*}[tp]
\centering
\caption{The results of $AP{50}$ and $AP_{75}$ on FSOD datasets. When testing, our setting is 5-shot and 2-way flowing FSOD sittings. We only train on training set and evaluate directly on test set. The best result is highlighted.}
\begin{tabular}{l|c|c|c|c}%l=left,r=right,c=center
\hline Model  &FSOD pretrain &  fine-tune &  AP$_{50}$ &  AP$_{75}$ \\
\hline FRCNN~\citep{Renetal2015} &no &yes &11.8 &6.7 \\
FRCNN~\citep{Renetal2015} &yes &yes &23.0 &12.9 \\
LSTD~\citep{ChenHetal2018} &yes &yes &24.2 &13.5 \\
FSOD~\citep{Fanetal2020} &yes &no &27.5 &19.4 \\
MM-FSOD(ours) &yes &no &{\bf 51.7} &{\bf 31.1} \\ \hline
\end{tabular}
\label{tab1}
\end{table*}

\subsubsection{Result on FSOD}
In Table \ref{tab1}, we compare our MM-FSOD with FSOD~\citep{Fanetal2020}, FRCNN~\citep{Renetal2015} and LSTD~\citep{ChenHetal2018} on FSOD dataset to show the amazing performance of our method. For a fair comparison, we follow FSOD settings with 2-way 5-shot. And we train MM-FSOD on training dataset of FSOD, then test on testing dataset with 200 categories without fine-tuning. From the highlighted results, it can be seen that our MM-FSOD achieves state-of-the-art on AP$_{50}$ and AP$_{75}$. Surprisingly, our model surpasses over FSOD for 24.2$\%$ on AP$_{50}$, and even surpasses FSOD over 11.7$\%$ in more strict metrics AP$_{75}$. What needs to be highlighted is that our model only uses ResNet34 as backbone. Comparing with most methods~\citep{Wuetal2020, Kangetal2019, Fanetal2020}, our model uses a smaller backbone which is more compact. Unlike class-specific approach, our model does not need to fine-tune on new categories samples and just needs to learn several steps from support images.

\begin{table*}[tp]
\centering
\caption{$AP_{50}$ on different novel split of VOC datasets. We evaluate MM-FSOD for 1/3/5/10 shot using ResNet34 as backbone. The best result is highlighted}
\resizebox{\textwidth}{25mm}{
\begin{tabular}{l|cccc|cccc|cccc}
\hline \multicolumn{1}{c|}{ }  &\multicolumn{4}{|c|}{split1} &\multicolumn{4}{|c|}{split2} & \multicolumn{4}{|c}{split3}\\
\hline Model/shot &1&3&5&10 &1&3&5&10 &1&3&5&10\\
\hline
FRCN+joint~\citep{Yanetal2019}      &2.7    &4.3    &11.8   &29         &1.9    &8.1  &9.9    &12.6         &5.2    &6.4    &6.4    &6.4\\
FRCN+ft~\citep{Yanetal2019}         &11.9   &29.0     &36.9   &36.9       &5.9    &23.4 &29.1   &28.8         &5.0      &18.1   &30.8   &43.4\\
FRCN+ft+full~\citep{Yanetal2019}    &13.8   &32.8   &41.5   &45.6       &7.9    &26.2 &31.6 &39.1           &9.8    &19.1   &35     &45.1\\
FSRW~\citep{Kangetal2019}            &14.8   &26.7   &33.9   &47.2       &15.7   &22.7 &30.1 &39.2		    &19.2   &25.7   &40.6   &41.3\\
Meta R-CNN~\citep{Yanetal2019}       &19.9	&35	    &45.7	&51.5		&10.4	&29.6	&34.8	&45.4		&14.3	&27.5	&41.2   &48.1\\
FSOD-KT~\citep{Kimetal2020}         &27.8	&46.2	&55.2	&56.8	    &19.8	&38.7	&38.9	&41.5		&29.5	&38.6	&43.8	&45.7\\
MPSR~\citep{Wuetal2020}            &41.7	&51.4	&55.2	&{\bf61.8}		&24.4	&39.2	&39.9	&47.8	    &{\bf35.6}	&42.3	&{\bf48.0}	    &{\bf49.7}\\
MM-FSOD(ours)             &{\bf50.0}		&{\bf55.9}	&{\bf57.9}	&{60.9}		&{\bf37.3}	&{\bf45.7}	&{\bf46.5}	&{\bf48.2}		&{\bf35.6}	&{\bf43.3}	&44.1	&45.4\\
\hline
\end{tabular}}
\label{tab2}
\end{table*}

\begin{table*}[tp]
\centering
\caption{Mean accuracy of the three splits for 1/3/5/10 shot}
\begin{tabular}{lclllclllclllclll}
\hline
\multicolumn{1}{c|}{model/shot}    & \multicolumn{4}{c}{1}    & \multicolumn{4}{c}{3}     & \multicolumn{4}{c}{5}     & \multicolumn{4}{c}{10}   \\ \hline
\multicolumn{1}{l|}{FRCNN+joint~\citep{Yanetal2019}}   & \multicolumn{4}{c}{3.26}  & \multicolumn{4}{c}{6.3}  & \multicolumn{4}{c}{9.4}   & \multicolumn{4}{c}{16}   \\
\multicolumn{1}{l|}{FRCNN+ft~\citep{Yanetal2019}}      & \multicolumn{4}{c}{7.6}   & \multicolumn{4}{c}{25.3}  & \multicolumn{4}{c}{32.3}  & \multicolumn{4}{c}{36.4} \\
\multicolumn{1}{l|}{FRCNN+ft+full~\citep{Yanetal2019}} & \multicolumn{4}{c}{10.5}  & \multicolumn{4}{c}{26.0}  & \multicolumn{4}{c}{36.0}    & \multicolumn{4}{c}{43.3} \\
\multicolumn{1}{l|}{FSRW~\citep{Kangetal2019}}          & \multicolumn{4}{c}{16.6}  & \multicolumn{4}{c}{32.5}  & \multicolumn{4}{c}{34.9}  & \multicolumn{4}{c}{42.6} \\
\multicolumn{1}{l|}{Meta R-cnn~\citep{Yanetal2019}}    & \multicolumn{4}{c}{14.9} & \multicolumn{4}{c}{37.0}    & \multicolumn{4}{c}{40.6}  & \multicolumn{4}{c}{48.3} \\
\multicolumn{1}{l|}{FSOD-KT~\citep{Kimetal2020}}       & \multicolumn{4}{c}{25.7}  & \multicolumn{4}{c}{41.7}  & \multicolumn{4}{c}{46.0}    & \multicolumn{4}{c}{48.0}   \\
\multicolumn{1}{l|}{MPSR~\citep{Wuetal2020}}          & \multicolumn{4}{c}{33.9}  & \multicolumn{4}{c}{44.3}  & \multicolumn{4}{c}{47.7}  & \multicolumn{4}{c}{{\bf53.1}} \\
\multicolumn{1}{l|}{MM-FSOD(ours)}           & \multicolumn{4}{c}{{\bf41.0}} & \multicolumn{4}{c}{{\bf48.3}} & \multicolumn{4}{c}{{\bf50.0}} & \multicolumn{4}{c}{51.5} \\ \hline
\end{tabular}
\label{tab3}
\end{table*}
\subsubsection{Results on PASCAL VOC}
PASCAL VOC dataset contains 20 categories. Most previous methods~\citep{Kangetal2019,Yanetal2019,Wuetal2020} divide dataset into three splits which consist of base/new categories. We delete classes of PASCAL VOC in FSOD and pre-train on FSOD training datasets, then we train the model on different splits. When testing, our model also does not fine-tune on new classes. Table~\ref{tab2} shows that our MM-FSOD achieves state-of-the-art with 1/3/5-shot settings on split1 and split2. It is clear that our model shows all-round strong performance than previous methods. In particular, when testing on split1 and split2, our model performs surprisingly in 1-shot task, which improves respectively by 8.3$\%$ on split1 and 12.9$\%$ on split2 compared to MPSR~\citep{Wuetal2020}. When K is set to 10, our model obtains a competitive result. And on split2, we get the best result 48.23$\%$. On split3, the hardest split, we get the best result in the low-shot situation and a competitive result on 5-shot and 10-shot. According to overall results, our method can learn more generalized knowledge in low-shot scenario. In different situations (1/3/5/10-shot), our method has the stable ability, which means it can resist the instability causing by support images bias during inference.
\par As shown in Table~\ref{tab4}, It is clear that our model achieves state-of-the-art results in 1-shot, 3-shot, and 5-shot object detection tasks with an average improvement of 7.1$\%$, 4$\%$ and 2.3$\%$ respectively. In 10-shot task, we also have a competitive result (51.5$\%$) which is a little lower than MPSR. However, our model is a single scale based model while MPSR is a multi-scale based model. Furthermore, comparing to Meta R-CNN and MPSR with ResNet-101, our MM-FSOD adopts smaller ResNet-34 as backbone, which will consume less computing resources. The results show that even if we use a smaller backbone, MM-FSOD still achieves state-of-the-art results in 1/3/5 shot object detection tasks.

\begin{table}[htbp]
\centering
\caption{Few-shot detection on MS COCO datasets with 10-shot.}
\begin{tabular}{l|c|l|l|l|c|l|l|l|c|l|l|l}
\hline
\multicolumn{1}{c|}{model/shot} & \multicolumn{2}{c|}{dataset} & \multicolumn{2}{c}{AP$_{50}$} & \multicolumn{2}{c}{AP$_{75}$}  & \multicolumn{2}{c}{AP}   \\
\hline Meta R-CNN~\citep{Yanetal2019} & \multicolumn{2}{c|}{COCO} & \multicolumn{2}{c}{19.1} & \multicolumn{2}{c}{6.6}   & \multicolumn{2}{c}{8.7}  \\
FSRW~\citep{Kangetal2019}   & \multicolumn{2}{c|}{COCO} & \multicolumn{2}{c}{12.3} & \multicolumn{2}{c}{7.6}   &\multicolumn{2}{c}{5.6}  \\
ONCE~\citep{Perez-Ruaetal2020} & \multicolumn{2}{c|}{COCO} & \multicolumn{2}{c}{N/A}  & \multicolumn{2}{c}{N/A}   &\multicolumn{2}{c}{5.1}  \\
MPSR~\citep{Wuetal2020} & \multicolumn{2}{c|}{COCO} & \multicolumn{2}{c}{17.9} & \multicolumn{2}{c}{9.7} &\multicolumn{2}{c}{9.8}  \\
FSOD~\citep{Fanetal2020} & \multicolumn{2}{c|}{COCO} & \multicolumn{2}{c}{{\bf20.4}} &\multicolumn{2}{c}{{\bf10.6}}  & \multicolumn{2}{c}{{\bf11.1}} \\
FSOD~\citep{Fanetal2020} & \multicolumn{2}{c|}{FSOD} & \multicolumn{2}{c}{{\bf31.3}} &\multicolumn{2}{c}{{\bf16.1}}  & \multicolumn{2}{c}{{\bf16.6}} \\ \hline
MM-FSOD(ours) & \multicolumn{2}{c|}{COCO} & \multicolumn{2}{c}{19.2} & \multicolumn{2}{c}{8.0} &\multicolumn{2}{c}{8.2} \\
MM-FSOD(ours) & \multicolumn{2}{c|}{FSOD} & \multicolumn{2}{c}{29.0} & \multicolumn{2}{c}{15.6} &\multicolumn{2}{c}{15.7} \\ \hline
\end{tabular}
\label{tab4}
\end{table}

\subsubsection{Results on MS COCO}
MS COCO~\citep{Linetal2014} is recognized as a hard object detection dataset, and it is a popular benchmark dataset in object detection task. For COCO dataset, most methods~\citep{Wuetal2020, Yanetal2019,Perez-Ruaetal2020, Fanetal2020,Kangetal2019} apply 10-shot or 30-shot settings, so we follow their protocol. Different from the settings of previous two datasets, we set the RPN NMS threshold from 0.7 to 0.5. Meanwhile, we decrease meta learning rate from 0.01 to 0.005 and increase inner loop steps to 50. Moreover, the score threshold is adjusted to 0.4. we adopt this setting because there is too much complex background information in COCO dataset, which is not current learning categories. As it is difficult to remove other irrelevant categories information from the support images, the predictions have lower confidence scores. Table~\ref{tab4} shows MM-FSOD's performance on 10-shot object detection task against state-of-the-art algorithoms. Compared to previous state-of-the-art result~\citep{Fanetal2020}, MM-FSOD achieves close performance. In few-shot object detection field, our method and \citep{Fanetal2020} surpass other methods by big margin.

\begin{figure}[ht]
\centering
\subfigure[]{
\includegraphics[width=5.5cm, height=4cm]{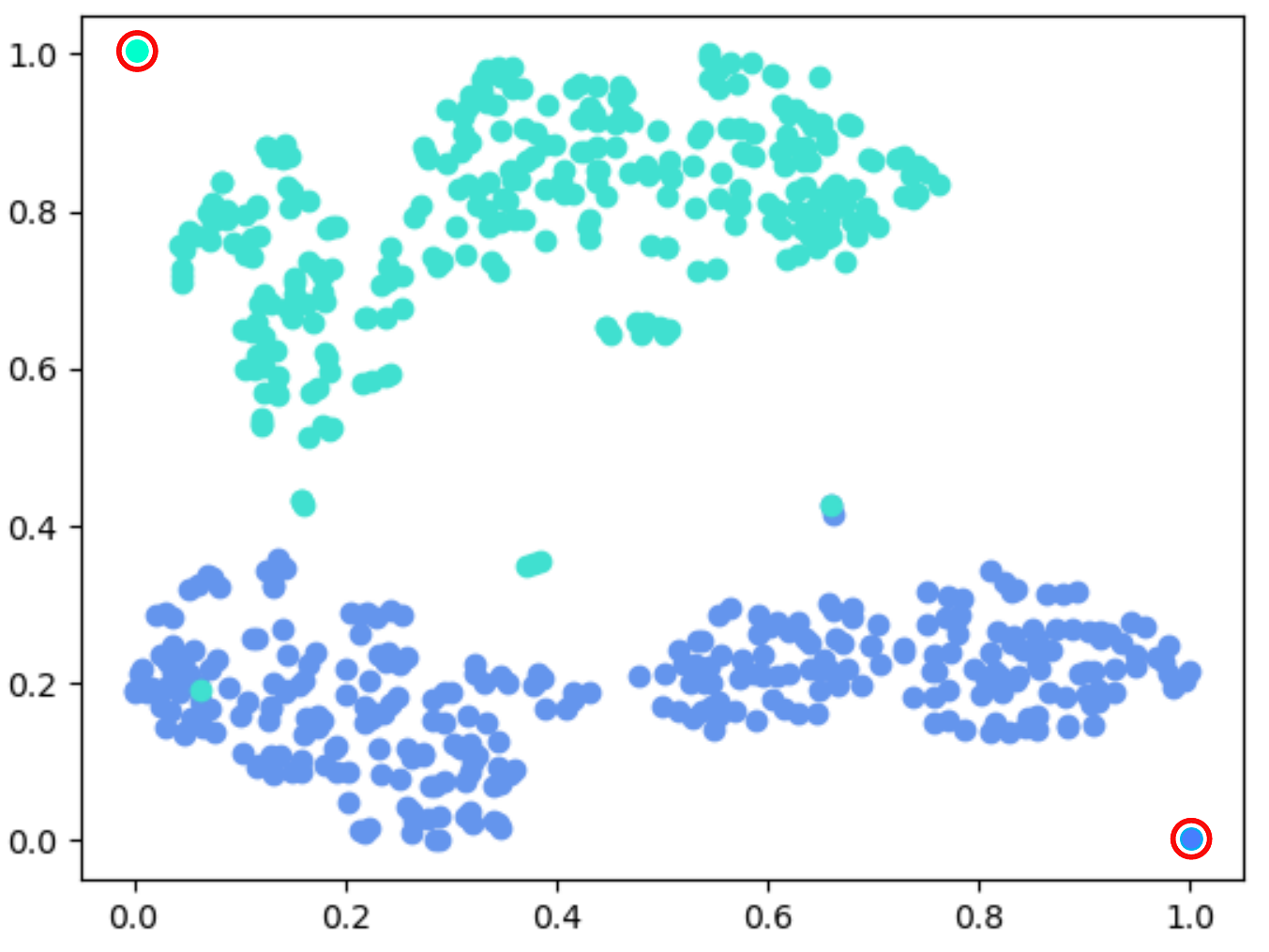}
\label{Fig6.a}
%\caption{fig1}
}
\quad
\subfigure[]{
\includegraphics[width=5.5cm,height=4cm]{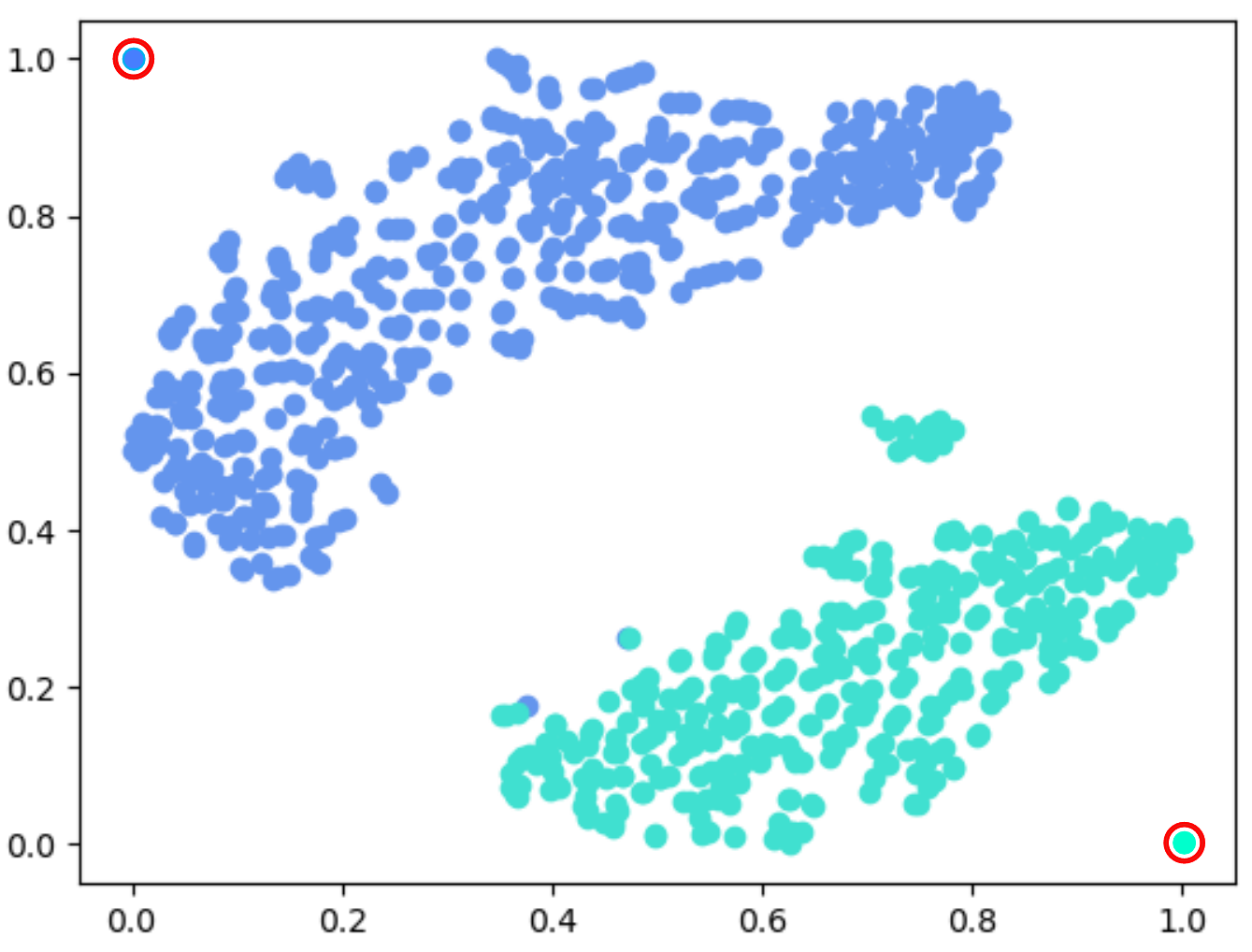}
\label{Fig6.b}
}

\subfigure[]{
\includegraphics[width=5.5cm,height=4cm]{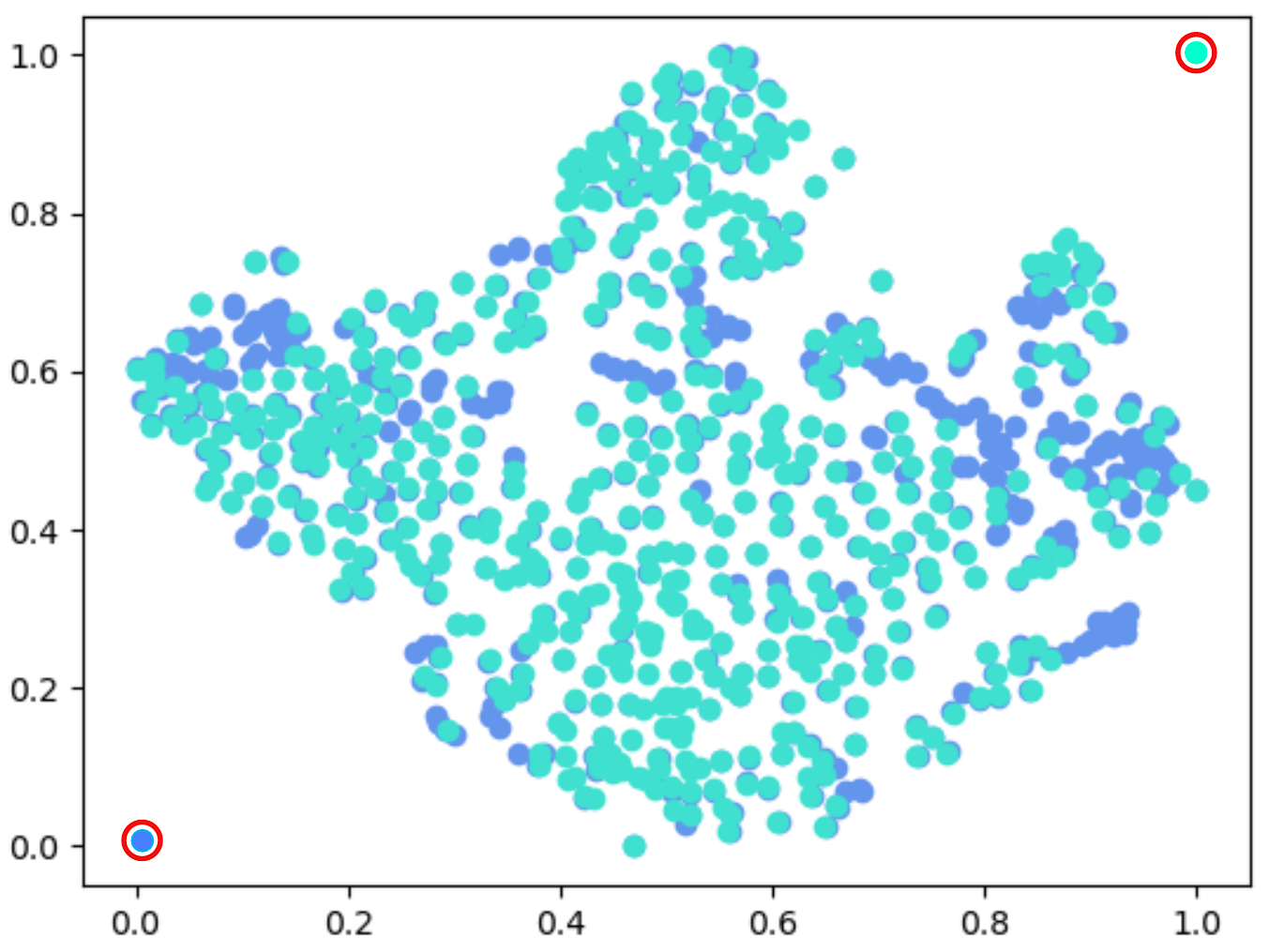}
\label{Fig6.c}
}
\quad
\subfigure[]{
\includegraphics[width=5.5cm,height=4cm]{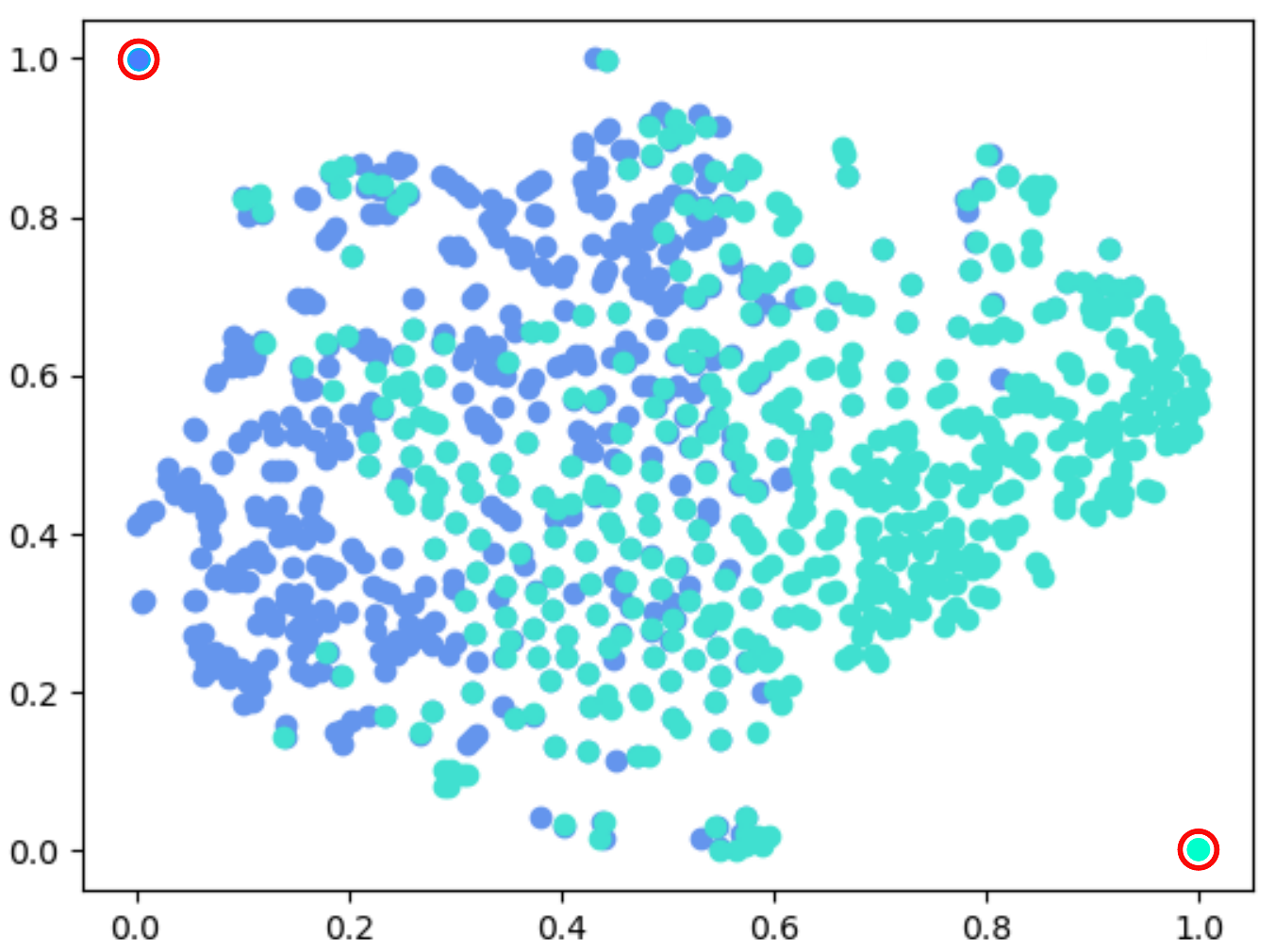}
\label{Fig6.d}
}
\caption{The RoI features' distribution conducted by T-SNE method. The Fig (a) and (b) shows two different images' reconstructed features of RoIs after MR module. Fig (c) and (d) shows same two images' RoIs features processed by Faster R-CNN ResNet. The dots in red cycle are prototypes of different classes. Other dots are query RoIs.}
\label{Fig6}
\end{figure}

\subsection{Analysis of results}
Previous results show that MM-FSOD obtain encouraging performance on different datasets. In this section, we conduct ablation experiments to verify the effectiveness of different modules in MM-FSOD. We mainly analyze the contribution of MR module and Pearson distance in MM-FSOD.

\subsubsection{Discussion on meta-representation module}
 We experimentally found that the backbone has the ability to extract low-level features of new classes. So it is not necessary to train the whole model by using the MAML's training method. Although the backbone can extract the features of new classes in shallow layers, it is difficult to express high-level features, which causes the decline of classification accuracy. To solve this problem, MR module is proposed by reconstructing the different categories features using intra-class mean prototypes~\citep{Snelletal2017}. This design can enlarge the inter-class distance in high-level embedding feature space.

\par From Table~\ref{tab5}, we can see that ``ResNet34 + PR'' is basically incapable of learning new categories without MR module. We train a Reset34 based Faster R-CNN with 2-way 5-shot settings similar to MM-FSOD. Specifically, ``ResNet34 + PR'' utilizes the support prototypes by taking the intra-class mean value of support features after ResNet34 layer4. And then we use Pearson distance(PR) to classify RoIs. When adding MR module, the model makes a significant contribution to represent low-level features, which leads to large improvement on detection results. The model ``ResNet34 + MR + PR'' brings 39.2\% and 24.1\% improvement than ``ResNet34 + PR'' respectively on FSOD and MS COCO datasets. With larger backbone (ResNet50), MM-FSOD performs even better. Especially on dataset FSOD, MM-FSOD gains improvement by large margin (11.5\%).
\begin{table}[tp]
\centering
\caption{Tesing on FSOD and MS COCO with different module. The metric is $AP_{50}$}
\begin{tabular}{l|c|c}
\hline
\multicolumn{1}{c|}{model/dataset} & FSOD & COCO \\ \hline
ResNet34 + PR                           & 12.5 & 3.4 \\
ResNet34 + MR + PR       & 51.7 & 27.5 \\
ResNet50 + MR + PR       & {\bf63.2} & {\bf29.0} \\
\hline
\end{tabular}
\label{tab5}
\end{table}

\par As can be seen in Fig~\ref{Fig6}, we use T-SNE~\citep{Maatenetal2008} to reduce the dimension of support prototypes(the dot in the red circle) extracted by the MR module and normalize the data for displaying in 2D plane. Fig~\ref{Fig6} clearly shows that MR module can divide the intra-class mean prototypes of each class into separate groups. Large distances for prototypes of different classes can enhance discrimination. Fig.\ref{Fig6.a} and Fig.\ref{Fig6.b} shows the distribution of $\mathbf{v_{meta}}$ (Eq.\ref{eq2}) on two different input images. Fig.\ref{Fig6.c} and Fig.\ref{Fig6.d} show the comparison of ResNet34 based Faster R-CNN. Fig.\ref{Fig6.a} and Fig.\ref{Fig6.b} show that the distribution of each query RoI feature is clearly demarcated after feature reconstructing by MR module. And all of them are distributed towards their intra-class mean prototype region respectively. Fig.\ref{Fig6.c} and Fig.\ref{Fig6.d} show that without MR module, the distribution of each class's RoIs features will not cluster around their intra-class mean prototype. This phenomenon can explain why Faster R-CNN without MR module is incapable of detecting the new classes. According to Tab~\ref{tab5} and Fig.~\ref{Fig6}, MR module has the ability to reconstruct the features of different categories after learning from support intra-class mean prototypes. Furthermore, MR module increases the distance of different categories of RoI features.

\begin{table}[ht]
\centering
\caption{Tesing cosine distance and Pearson distance on FSOD with 5-shot and 10-shot on COCO datasets.}
\begin{tabular}{l|c|c}
\hline
\multicolumn{1}{c|}{model/dataset} & FSOD & COCO \\ \hline
ResNet34 + PR                           & 12.5 & 3.4 \\
ResNet34 + MR + CD       & 33.7 & 10.9 \\
ResNet34 + MR + PR       & {\bf51.7} & {\bf27.5} \\
 \hline
\end{tabular}
\label{tab6}
\end{table}

\subsubsection{Discussion on Pearson distance}
Currently, in the field of metric learning, most methods~\citep{LiuJetal2019, LiWetal2019} use cosine operation to measure distance of different classes. And ohter approaches~\citep{ZhangCetal2020, BateniPetal2020} design different distance metric to measure the similarity of different classes. Our approach uses Pearson distance to compute the similarity of different features in feature space. We adopt Pearson distace according to the following princeples: (1) MR module is used to reconstruct the features of query RoIs by taking the intra-class mean of different classes features as representative prototypes for classification; (2) The Pearson distance calculates the similarity between two features, which takes into account the effects of the variance causing from complex background information. Therefore, adopting the Pearson distance is more suitable for our model. Experiments also prove the efficiency of Pearson distance. Table \ref{tab6} shows that our model achieves better results with Pearson distance by nearly 17$\%$.

\par It can be seen in Table \ref{tab6}, all three experiments use ResNet34 as the base feature extraction module. PR represents the metric module using Pearson distance, where CD represents the metric module using cosine distance. In 2-way 5-shot experiment, compared to ``ResNet34 + MR + CD'', the model ``ResNet34 + MR + PR'' with Pearson distance has significantly higher accuracy than the model with cosine distance on AP$_{50}$. The model that uses the Pearson distance improved by 18$\%$ on the FSOD dataset and by 16.6$\%$ on the COCO dataset compared to model with cosine distance. This result shows that the MR module with Pearson diatance successfully improves the performance of our model in few-shot object detection task.

\subsubsection{Results visualization}
\begin{figure*}[tp]
  \centering
  \includegraphics[scale=0.12]{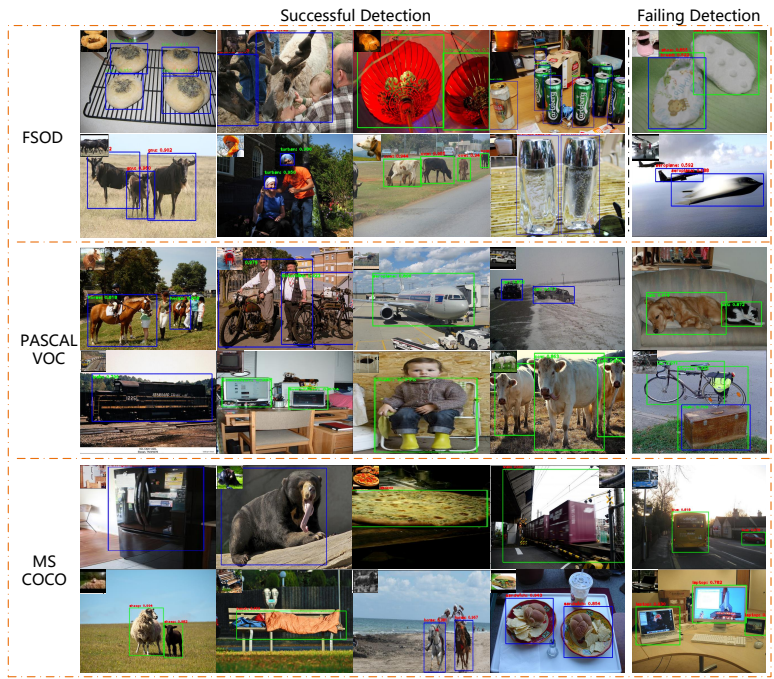}
  \caption{Visualization of detected results on FSOD, MS COCO and PASCAL VOC datasets. The experiment is conducted under 2-way 5-shot setting. The top corner of every image is the examples of support images.}
  \label{Fig7}
\end{figure*}

Fig.~\ref{Fig7} shows the MM-FSOD detection results, where the top left corner or right corner of each image is support samples. We use 2-way 5-shot settings, then we perform inference experiments on the test sets of FSOD, MS COCO, and PASCAL VOC with ResNet34 based model, respectively. According to the results, MM-FSOD shows a satisfactory performance. Most important, it can quickly learn the novel class and achieve accurate detection without fine-tuning. Of course, our model still misses the detection and makes false predictions when there are large disturbances in the support data. Overall, the performance of our proposed MM-FSOD is satisfactory. And it is a suitable approach to tackle few-shot object detection tasks.
\section{Conclusion}
We propose a novel and efficient few-shot object detection framework (MM-FSOD) which combines meta-learning and metric learning to tackle few-shot object detection tasks. MM-FSOD's innovations can be summarized as follows. Firstly, we transfer meta learning training method from classification to feature reconstruction. The MR module learns to classify the intra-class mean prototypes and then reconstruct low-level features. Secondly, we introduce Pearson distance into metric learning module and prove that it is more effective than cosine distance. Last but not least, we apply the metric learning similarity function into object detection and integrate it with the meta-learning approach to achieve better detection performance. To our knowledge, it is the first class-agnostic object detection model combining these two learning methods. The results of different datasets show that our MM-FSOD model achieves amazing performance and realizes fast convergence on novel classes.

\section*{Acknowledgments}
This work was supported by the National Natural Science Foundation of China(Grant NO.62072021).

\bibliography{mybibfile}

\end{document}